\documentclass{article}

\usepackage{arxiv}

\usepackage[utf8]{inputenc} 
\usepackage[T1]{fontenc}    
\usepackage{hyperref}       
\usepackage{url}            
\usepackage{booktabs}       
\usepackage{amsfonts}       
\usepackage{nicefrac}       
\usepackage{microtype}      
\usepackage{lipsum}		
\usepackage{graphicx}
\usepackage{natbib}
\usepackage{doi}
\usepackage{algorithm}
\usepackage{algorithmic}
\usepackage{amsmath}

\usepackage{appendix}

\usepackage{ccaption}
\usepackage{booktabs}

\usepackage{fancyhdr}

\usepackage{multirow}
\usepackage{amsmath}
\usepackage{amssymb}
\usepackage{mathrsfs}

\title{Feature Aligning Few shot Learning Method Using Local Descriptors Weighted Rules}

\author{ \href{https://orcid.org/0000-0000-0000-0000}{\includegraphics[scale=0.06]{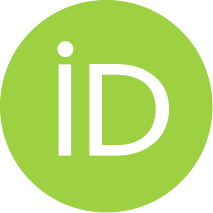}\hspace{1mm}Bingchen Yan}\thanks{} \\
	Department of Computer Science\\
	Guilin University Of Electronic Technology\\
	Guilin, Gungxi, China  \\
	\texttt{1321847667a@gmail.com} \\
}

\renewcommand{\shorttitle}{\textit{arXiv} Template}

\hypersetup{
pdftitle={Feature Aligning Few shot Learning Method Using Local Descriptors Weighted Rules},
pdfsubject={q-bio.NC, q-bio.QM},
pdfauthor={Bingchen Yan},
pdfkeywords={First keyword, Second keyword, More},
}

\begin{document}
\maketitle

\begin{abstract}
	
 Few-shot classification involves identifying new categories using a limited number of labeled samples. Current few-shot classification methods based on local descriptors primarily leverage underlying consistent features across visible and invisible classes, facing challenges including redundant neighboring information, noisy representations, and limited interpretability. This paper proposes a Feature Aligning Few-shot Learning Method Using Local Descriptors Weighted Rules (FAFD-LDWR). It innovatively introduces a cross-normalization method into few-shot image classification to preserve the discriminative information of local descriptors as much as possible; and enhances classification performance by aligning key local descriptors of support and query sets to remove background noise. FAFD-LDWR performs excellently on three benchmark datasets , outperforming state-of-the-art methods in both 1-shot and 5-shot settings. The designed visualization experiments also demonstrate FAFD-LDWR's improvement in prediction interpretability.
\end{abstract}


\section{Introduction}
Deep learning models have achieved significant success in various computer vision domains with large-scale annotated datasets \cite{wang2023optimizing,wang2024fast,zhou2024spike}. However, they struggle with new classes containing only a few labeled samples, often leading to overfitting or failure to converge. In contrast, humans can recognize new classes from a few examples by leveraging prior knowledge. Few-shot learning addresses this by generalizing knowledge from base classes (with abundant samples) to novel classes (with few samples), attracting increasing attention. Effective few-shot learning methods can be broadly categorized into metric-based \cite{zheng2023bdla,sun2024klsanet,snell2017prototypical,vinyals2016matching,li2019distribution,li2020more,huang2021local,qi2022task,sung2018learning}, meta-learning-based \cite{leng2024meta,finn2017model,lee2019meta}, and transfer-based \cite{sun2021explanation,chen2021meta,fu2021meta,tseng2020cross,gao2024few,hu2022adversarial} approaches. Notably, metric-based methods have achieved great success due to their simplicity and effectiveness. This paper focuses on metric-based methods, which typically involve: 1) extracting features from query and support images; 2) computing distances between the query image and each support image, prototype, or class center; and 3) assigning labels to the query image through nearest neighbor search.

Despite their success, metric-based methods are often troubled by noise from irrelevant local regions \cite{chen2024featwalk,zheng2023bdla,sun2024klsanet,zhou2024global}, as the semantic content of local areas can vary significantly. As illustrated in Figure 1, some regions contain key semantics consistent with the image class (e.g., the "bird" region in a "bird" image), while others may contain irrelevant semantics (e.g., the "tree" region in a "bird" image).

To address this issue, GLIML \cite{hao2021global}and KLSANet \cite{sun2024klsanet}use a dual-branch architecture to learn both global and local features, selecting local features based on their similarity to global features. Although effective, this approach increases model complexity and runtime. BDLA \cite{zheng2023bdla}proposes calculating bidirectional distances between the local features of query and support samples to enhance semantic alignment.
\begin{figure}[ht]
\centering
\includegraphics[scale=0.4]{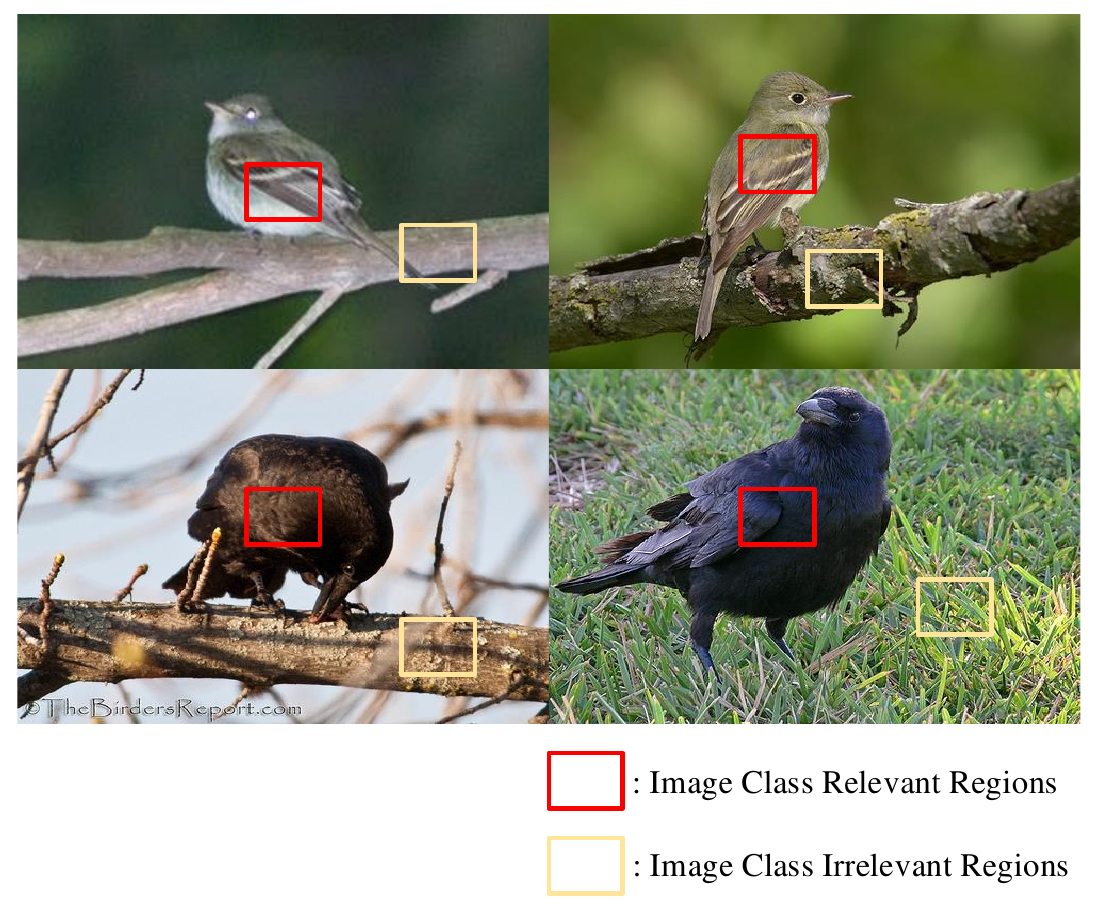}
\caption{Examples Of Regions That Are Relevant And Irrelevant To Image Classes.}
\label{region}
\end{figure}
Building on previous work, our method uses local descriptor-level features to eliminate noise regions irrelevant to the image class. We propose a novel few-shot learning method based on dynamically weighted local descriptor filtering. Experimental results on three commonly used few-shot learning datasets surpass current state-of-the-art methods. Remarkably, our method even outperforms recent transfer learning-based few-shot learning methods on the CUB-200 dataset, suggesting significant implications for future research in few-shot learning.

Our contributions are as follows:
\begin{itemize}
\item We innovatively introduce the cross-normalization method into few-shot learning, preserving the discriminative information of local descriptor features.
\item  We propose using the neighborhood representation of local descriptors instead of directly using the local descriptors. This approach not only utilizes the information of individual local descriptors but also incorporates the contextual information of their neighborhoods. Calculating the mean of neighbors as a new representation can smooth out local noise, enhancing feature stability and robustness. This effectively addresses the limitation of solely relying on local descriptors, which may overlook surrounding context, making our feature representation more comprehensive and representative.
\item  We propose a dynamic method for filtering out local descriptors irrelevant to class information, thereby improving classification performance.
\end{itemize}
If the paper is accepted our code and experimental data will be made public.

\section{Related work}
Current approaches to few-shot image classification can be broadly categorized into two main strategies: optimization-based techniques and Metric-learning based methodologies.

Optimization-based approaches, often referred to as meta-learning, seek to establish a robust starting point for model parameters. This initialization encapsulates prior knowledge and experience, enabling swift adaptation to new tasks through minimal gradient updates. Examples of this approach include techniques like MAML \cite{finn2017model}, MetaOptNet \cite{lee2019meta}, and FIAML-LR \cite{wang2024fast}.
On the other hand, Metric-learning based methods focus on learning a function to gauge the likeness between samples for classification or regression tasks. For instance, Prototypical Networks \cite{snell2017prototypical}create a representative `prototype' for each class in the feature space, classifying new samples based on their proximity to these prototypes. Matching Networks \cite{vinyals2016matching}employ a weighted sum of similarities between a query sample and the support set to determine class membership.

Our research primarily explores Metric-learning based techniques. Notable work in this area includes DN4 \cite{li2019revisiting} innovative use of metrics at the local descriptor level. This approach mitigates the loss of discriminative information that can occur when condensing an image's local features into a single, compact representation. Their method calculates the k-nearest local features from query examples to support examples, yielding impressive results. Building on this, BDLA \cite{zheng2023bdla}introduced bidirectional distance calculations between query and support samples, enhancing the alignment of contextual semantic information.
\subsection{Application Of Local Descriptors In Few-Shot Learning}
Local descriptors are crucial in many computer vision tasks and can be broadly categorized into patch-based and dense descriptor methods. Patch-based methods, such as L2Net \cite{tian2017l2}, HardNet \cite{mishchuk2017working}, SOSNet \cite{tian2019sosnet}, and ContextDesc \cite{luo2019contextdesc}, extract local descriptors from each image patch. Dense descriptor methods, including SuperPoint \cite{detone2018superpoint}, D2Net \cite{dusmanu2019d2}, R2D2 \cite{revaud2019r2d2}, CAPS \cite{wang2020learning}, ASLFeat \cite{luo2020aslfeat}, and DGDNet \cite{liu2021dgd}, use fully convolutional neural networks \cite{long2015fully} to extract dense local descriptors from the entire image.

Recent works incorporating local descriptors into few-shot learning have shown remarkable effectiveness \cite{huang2021local,li2019revisiting,li2020more,qi2022task,sung2018learning}. For instance, LMP-Net \cite{huang2021local}addresses the limitation of prototype networks that use global features to calculate a single class prototype by employing local descriptor-level features to learn multiple prototypes per class, thus representing the class distribution more comprehensively. DN4 \cite{li2019revisiting} uses local descriptor representation and measures the relationship between images and classes by calculating the similarity between local descriptors with $k$-nearest neighbors ($k$-NN) as the classifier. Similarly, the Relation Network \cite{sung2018learning} implicitly measures the distance between query and support samples using local descriptors.

However, treating all local descriptors indiscriminately overlooks two potential drawbacks in few-shot image classification: first, local descriptors often contain redundant background information that is not valuable for classification; second, semantically shared local descriptors across classes are not crucial for recognizing novel instances.

\begin{figure*}[ht]
\centering
\includegraphics[scale=0.8]{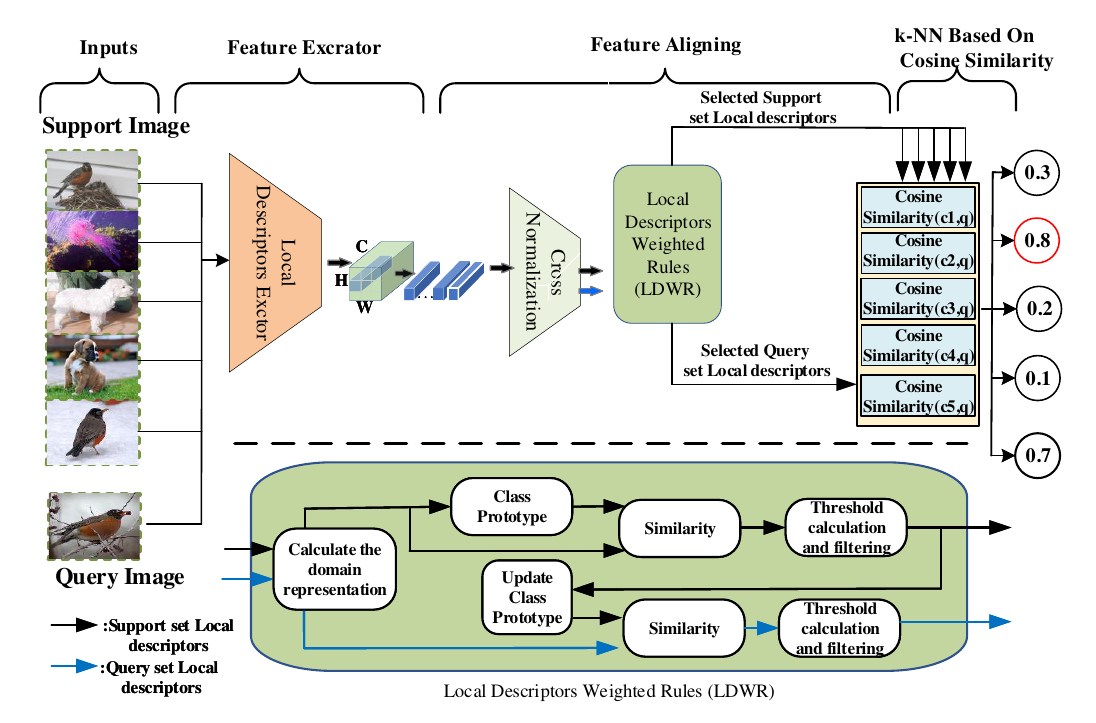}
\caption{The proposed FAFD-LDWR method's framework for 5-way 1-shot classification.}
\label{main}
\end{figure*}

\section{ Method}
As shown in Figure \ref{main}, our FAFD-LDWRM method comprises three main components: the embedding feature extraction module, the cross normalization module, and the local descriptors with dynamically weighted rules module.

First, the embedding feature extraction module uses an embedding network based on a contextual learning mechanism to extract features from the support and query set images. Second, the cross normalization module normalizes the spatial and channel dimensions of the local descriptors using adaptive parameters, retaining maximum discriminative information. Finally, the local descriptors with dynamically weighted rules module calculates the weight of each local descriptor, filters out key descriptors, removes background noise, and enhances few-shot classification performance.

\subsection{Problem Formulation}
Few-shot learning focuses on enabling models to perform well with a minimal number of samples while ensuring strong generalization capabilities. Specifically, we address the $N$-way $K$-shot problem, where $N$ indicates the number of classes and $K$ represents the number of samples per class. Typically, $K$ is a small number, such as 1 or 5.

Given a training dataset \(D^{train} = \{(x_r, y_r)\}^T_{r=1}\), the task T of few-shot learning aims at learning the model parameters $\theta$ that allow for quick adaptation to an unseen test dataset \(D^{test}\) using an episodic training mechanism \cite{vinyals2016matching}. Here, each \(y_r\) denotes the true label of the image \(x_r\). In both \(D^{train}\) and \(D^{test}\), each episode comprises a support set \(S\) and a query set \(Q\). The support set \(S\) consists of $N$ distinct image classes, each containing $K$ randomly labeled images. The query set \(Q\) is utilized for evaluating the model.

\subsection{Embedding Feature Extraction Module}
Following previous work,we utilize a Conv-4 or Resnet-12 to serve as a local descriptor feature extractor. When we process an image \( I \) through local descriptor feature extractor, the output is a three-dimensional array \( \mathcal{A}_\phi(I) \in \mathbf{R}^{C \times H \times W} \). This array encapsulates the image's characteristics, where \( \mathcal{A}_\phi(I) \) denotes the transformation learned by the neural architecture, \( \phi \) represents the neural network's parameters, and \( C \), \( H \), \( W \) signify the array's channels, height and width, respectively. We can express this mathematically as:

\begin{equation}
    \mathcal{A}_\phi(I) = [\mathbf{x}^1, ..., \mathbf{x}^N] \in \mathbf{R}^{C \times N}
\end{equation}

In this formulation, \( N = H \times W \), which maps all images to a common representational domain. Each three-dimensional array comprises \( N \) units of \( C \) dimensions, with each unit embodying a localized feature of the image. Compared to single-dimensional \cite{vinyals2016matching,snell2017prototypical} or alternative dimensional representations, three-dimensional arrays more effectively preserve spatial relationships\cite{zheng2023bdla}. Consequently, in the context of few-shot learning within similarity-based frameworks, three-dimensional arrays are frequently preferred. In our study, we utilize these three-dimensional array features to represent both the support set S and query set Q.

\subsection{Cross Normalization Module}
Unlike existing few-shot learning methods that focus on normalizing local descriptors using L2 normalization, our approach is inspired by the success of cross normalization in tasks such as image matching, homography estimation, 3D reconstruction , and visual localization \cite{wang2022cndesc}. Our method emphasizes normalizing and summing the spatial and channel dimensions of local descriptors through adaptive parameters to retain as much discriminative information as possible.

First, we perform spatial-level normalization on the local descriptors:
\begin{equation}
\begin{aligned}
x_s = \left( \frac{x - \mu}{\sqrt{\sigma^2 + \epsilon}} \right) \cdot \text{conv1}(\text{map}) + \text{conv2}(\text{map})
\end{aligned}
\end{equation}
where \(x\) is the input feature, \(\mu\) and \(\sigma^2\) are the computed mean and variance, respectively, and \(\epsilon\) is a small constant to avoid division by zero errors. To further enhance the features, we process the normalized mean map through two parallel \(1 \times 1\) convolution layers, denoted as `conv1' and `conv2'.

Next, we apply channel-level normalization to independently normalize each channel of the local descriptors:

\begin{equation}
x_c = \gamma \times \frac{x - \mu}{\sqrt{\sigma^2 + \epsilon}} + \beta
\end{equation}

Similar to spatial-level normalization, \(\mu\) and \(\sigma^2\) are the mean and variance computed along the spatial dimensions (height and width). To make the normalization process more flexible and effective, we introduce two adaptive learnable parameters, \(\gamma\) and \(\beta\), which dynamically adjust the scale and offset based on the input local descriptor features.

Finally, we adopt a feature fusion strategy to combine the results of spatial-level normalization and channel-level normalization by weighted fusion, enhancing the model's ability to discriminate local features in query images and support sets. The specific fusion process is as follows:
\begin{equation}
x_{CN} = x_s \times \frac{\omega_1}{\omega_1 + \omega_2} + x_c \times \frac{\omega_2}{\omega_1 + \omega_2}
\end{equation}

where \(\omega_1\) and \(\omega_2\) are learnable fusion weights, and \(x_s\) and \(x_c\) are the outputs of the two normalization strategies, respectively. The details of cross normalization can be  found in Supplementary Sections A.

Through this advanced normalization and adaptive parameter adjustment strategy, we not only retain the discriminative information of the features but also improve the model's performance in few-shot image recognition tasks, demonstrating superior results compared to traditional L2 normalization methods.

\subsection{Local Descriptors With Dynamically Weighted Rules Module}

We propose a novel strategy to evaluate local descriptor importance, aiming to filter and retain key descriptors. This improves classification accuracy by preventing the classifier from learning irrelevant features.

Our strategy is based on an observation: local descriptor neighborhoods often exhibit consistent visual patterns within image categories. Background descriptors tend to be similar across images, while main subject descriptors share similarities within their category. We describe this as ``one is influenced by those around them."

Based on this insight, we propose the following method to utilize the neighborhood information of local descriptors:

For each local descriptor, we use a $k$-NN algorithm based on cosine similarity to find its k most similar neighbors. First, we calculate the cosine similarity between the local descriptor \( q \) and all other local descriptors \( x_i \):

\begin{equation}
\begin{aligned}
\text{similarity}(q, x_i) = \frac{q \cdot x_i}{|q| |x_i|}
\end{aligned}
\end{equation}
Then, we select the top k local descriptors with the highest similarity as neighbors. This process can be represented as:

\begin{equation}
\begin{aligned}
\text{NN}_k(q) = \text{argtop}_k(\text{similarity}(q, x_i))
\end{aligned}
\end{equation}

where \( \text{argtop}_k \) denotes selecting the indices of the top k similarities. After obtaining the k nearest neighbors, we compute their mean to represent the neighborhood of the local descriptor \( q \):
\begin{equation}
\begin{aligned}
N_q = \frac{1}{k} \sum_{i \in \text{NN}_k(q)} x_i
\end{aligned}
\end{equation}
This process is performed for each local descriptor to obtain their respective neighborhood representations:

\begin{equation}
\begin{aligned}
N_i = \frac{1}{k} \sum_{j=1}^{k} x_{j},
\end{aligned}
\end{equation}
where \( x_j \) denotes the j-th nearest neighbor local descriptor.

This approach integrates both individual local descriptors and their contextual information. Computing neighborhood means smooths local noise, enhancing feature stability and robustness. It compensates for the limitations of relying solely on individual descriptors, resulting in a more comprehensive and representative feature representation.

Inspired by the idea of Prototypical Networks \cite{snell2017prototypical}, we compute the class prototype \( P_c \) for each support set category by averaging the local descriptor features. This class prototype encompasses more comprehensive and representative information related to the support set category, which is used for filtering key local descriptors. Specifically, the class prototype \( P_c \) is computed as follows:
\begin{equation}
\begin{aligned}
P_c = \frac{1}{|S_c|} \sum_{x \in S_c} f_{\theta}(x),
\end{aligned}
\end{equation}

where \( S_c \) denotes the support set of category \( c \), and \( f_{\theta}(x) \) represents the local descriptor-level feature embedding of sample \( x \).

Next, we calculate the cosine similarity \( S_{i,c} \) between the neighborhood representation \( N_i \) and the class prototypes \( P_c \) of the five support set categories:

\begin{equation}
\begin{aligned}
S_{i,c} = \frac{N_i \cdot P_c}{\|N_i\| \|P_c\|},
\end{aligned}
\end{equation}

where \( \cdot \) denotes the dot product, and \( \|\cdot\| \) represents the L2 norm of the vector.

\subsubsection{Weight aggregation and expansion}

The similarity of the neighborhood representation of each local descriptor to the support set categories consists of five similarities, each corresponding to a specific category. We average these five similarities to determine the importance of the neighborhood representation of the local descriptor across the five categories, indicating whether the local descriptor is important in the main subject of images from these categories.

The formula is as follows:
\begin{equation}
\begin{aligned}
\overline{\omega}_{i} = \frac{1}{K} \sum_{c=1}^{K} S_{i,c}
\end{aligned}
\end{equation}

where \(\overline{\omega}_{i}\) represents the average weight of the \(i\)-th local descriptor, \(K\) denotes the number of categories, and \(S_{i,c}\) represents the weight of the neighborhood representation of the \(i\)-th local descriptor for the \(c\)-th category, which is the cosine similarity from the previous step.

Using this formula, we obtain a weight matrix  with the shape \([M, K, T]\), where \(M\) denotes the number of samples in the support or query set, \(K\) denotes the number of categories, and \(T\) denotes the number of local features per sample.

To determine the adaptive threshold, we first calculate the mean and standard deviation of the average weights of all local descriptor neighborhood representations:
\begin{equation}
\begin{aligned}
\overline{\mu} = \frac{1}{M \times T} \sum_{m=1}^{M} \sum_{n=1}^{T} \overline{\omega}_{n,m}
\end{aligned}
\end{equation}

\begin{equation}
\begin{aligned}
\overline{\sigma} = \sqrt{\frac{1}{M \times T} \sum_{m=1}^{M} \sum_{n=1}^{T} (\overline{\omega}_{n,m} - \overline{\mu})^2}
\end{aligned}
\end{equation}

where \(\overline{\mu}\) represents the mean of the average weights of all local descriptor neighborhood representations, \(\overline{\sigma}\) represents the standard deviation of the average weights, and \(\overline{\omega}_{n,m}\) represents the average weight of the \(n\)-th local descriptor of the \(m\)-th support or query sample.

The filtering process involves two main steps. First, we calculate the cosine similarity scores $S_{i,c}$ between the neighborhood representation $N_i$ of each local descriptor in the support samples and the class prototype $P_c$. Next, we perform the Shapiro-Wilk test \cite{hanusz2016shapiro} on these cosine similarity scores and find that they approximately follow a normal distribution. Consequently, we apply the 3$\sigma$ principle, filtering out local descriptors with cosine similarity scores $S_{i,c}$ less than $\overline{\mu} - \overline{\sigma}$ as background descriptors irrelevant to the category.

The specific filtering formula is as follows:

\begin{equation}
\begin{aligned}
S_{i,c} < (\overline{\mu} - \overline{\sigma})
\end{aligned}
\end{equation}

We denote the standard deviation of the unfiltered cosine similarity scores as 
$\overline{\sigma_0}$ and iterate the above steps until $\overline{\sigma}$ is less than $\overline{\sigma_0} / C$, where $C$ is a predefined constant.

After filtering support set descriptors, we recompute class prototypes and apply the same filtering to the query set (see Algorithm 1 in Supplementary  for details).

\subsubsection{Classification using Selected Local Descriptors}

To classify the query image, we propose an improved image-to-class measurement method based on filtered local descriptors. This method fully leverages the filtered key local descriptors, effectively enhancing classification accuracy.

Specifically, given a query image \( q \), we first obtain its filtered local descriptor representation through our filtering mechanism:

\begin{equation}
\begin{aligned}
\mathcal{LDWR}_{\phi_{\text{filtered}}}(X_q) = [\hat{\mathbf{x}}^1_q, \hat{\mathbf{x}}^2_q, \ldots, \hat{\mathbf{x}}^L_q] \in \mathbb{R}^{C \times L}
\end{aligned}
\end{equation}

where \( L \leq N \) denotes the number of retained local descriptors after filtering. Similarly, each category \( i \) ( \( i = 1, 2, \ldots, 5 \) ) in the support set undergoes local descriptor filtering.

For each filtered key local descriptor of the query image $q$, we find its $\overline{k}$ nearest neighbors among the filtered local descriptors of each support category, denoted as $m_1, m_2, \ldots, m_{\overline{k}}$, and compute the corresponding cosine similarities: $\cos(\hat{\mathbf{x}}^1_q, m_1), \cos(\hat{\mathbf{x}}^2_q, m_2), \ldots, \cos(\hat{\mathbf{x}}^L_q, m_{\overline{k}})$.

Based on this, we define the similarity score between the query image \( q \) and category \( i \) as:

\begin{equation}
\begin{aligned}
\text{Similarity}(q, \text{category}_i) = \sum_{l=1}^L \sum_{j=1}^k \cos(\hat{\mathbf{x}}^l_q, m^i_j)
\end{aligned}
\end{equation}

Subsequently, we apply the softmax function to obtain the probability that the query image \( q \) belongs to category \( i \):

\begin{equation}
\begin{aligned}
P(c=i | q) = \frac{\exp(\text{Similarity}(q, \text{category}_i))}{\sum_{i=1}^5 \exp(\text{Similarity}(q, \text{category}_i))}
\end{aligned}
\end{equation}

This improved method not only fully utilizes the filtered key local descriptors but also enhances classification robustness by considering multiple nearest neighbors. By focusing on the most distinctive features within the image, our method can more accurately capture category-related information, thereby improving classification performance.

\section{Experiment}
\subsection{Datasets}
In this paper, we validate the effectiveness of our method using three commonly used few-shot classification benchmark datasets: CUB-200 \cite{welinder2010caltech}, Stanford Dogs \cite{khosla2011novel}, and Stanford Cars \cite{khosla2011novel}.The detailed introduction of datasets are presented in Supplementary Sections C.
\subsection{Experimental Setting}

In our experiments, we primarily focus on 5-way 1-shot and 5-shot classification tasks. To ensure fair comparison with other methods, we employ two commonly used backbone network structures in few-shot learning: Conv4 and ResNet-12, following the implementation details outlined in DN4 \cite{li2019revisiting}and CovaMNet \cite{li2019distribution}.The detailed experimental settings are presented in Supplementary Sections B.
\begin{table*}
    \caption{Comparison with STATE-of-the-art methods in the 5-way 1-shot and 5-shot settings.}
    \label{tab:results1}
    \centering
    \resizebox{\textwidth}{!}{
    \begin{tabular}{llcccccc}
        \toprule
        \multirow{2}{*}{Model} & \multirow{2}{*}{Embedding} & \multicolumn{2}{c}{Stanford Dogs} & \multicolumn{2}{c}{Stanford Cars} & \multicolumn{2}{c}{CUB-200} \\
        \cmidrule{3-8}
         &  & 1-shot & 5-shot & 1-shot & 5-shot & 1-shot & 5-shot \\
        \midrule

        Matching Net \cite{vinyals2016matching} & Conv4-64 & 35. 80$\pm$0. 99 & 47. 50$\pm$1. 03 & 34. 80$\pm$0. 98 & 44. 70$\pm$1. 03 & 45. 30$\pm$1. 03 & 59. 50$\pm$1. 01 \\
        Prototype Net \cite{snell2017prototypical} & Conv4-64 & 37. 59$\pm$1. 00 & 48. 19$\pm$1. 03 & 40. 90$\pm$1. 01 & 52. 93$\pm$1. 03 & 37. 36$\pm$1. 00 & 45. 28$\pm$1. 03 \\
        GNN \cite{fu2021meta} & Conv4-64 & 46. 98$\pm$0. 98 & 62. 27$\pm$0. 95 & 55. 85$\pm$0. 97 & 71. 25$\pm$0. 89 & 51. 83$\pm$0. 98 & 63. 69$\pm$0. 94 \\
        DN4 \cite{li2019revisiting} & Conv4-64 & 45. 41$\pm$0. 76 & 63. 51$\pm$0. 62 & 59. 84$\pm$0. 80 & 88. 65$\pm$0. 44 & 46. 84$\pm$0. 81 & 74. 92$\pm$0. 62 \\
        MADN4 \cite{li2020more} & Conv4-64 & 50. 42$\pm$0. 27 & 70. 75$\pm$0. 47 & 62. 89$\pm$0. 50 & 89. 25$\pm$0. 34 & 57. 11$\pm$0. 70 & 77. 83$\pm$0. 40 \\
        TDSNet \cite{qi2022task}& Conv4-64 & 52. 48$\pm$0. 87 & 66. 45$\pm$0. 49 & 57. 35$\pm$0. 91 & 73. 64$\pm$0. 72 & 67. 34$\pm$0. 85 & 79. 38$\pm$0. 59 \\
         BDLA \cite{zheng2023bdla} & Conv4-64 & 48. 53$\pm$0. 87 & 70. 07$\pm$0. 70 & 64. 41$\pm$0. 84 & 89. 04$\pm$0. 45 & 50. 59$\pm$0. 97 & 75. 36$\pm$0. 72 \\
        DLDA \cite{song2024learning} & Conv4-64 & 49. 44$\pm$0. 85 & 69. 36$\pm$0. 69 & 60. 86$\pm$0. 82 & \textbf{89. 50}$\pm$\textbf{0. 41} & 55. 12$\pm$0. 86 & 74. 46$\pm$0. 65 \\
        LCCRN \cite{li2023locally}& Conv4-64 & - & - & \textbf{71. 62}$\pm$\textbf{0. 21} & 86. 41$\pm$0. 12 & \textbf{76. 22}$\pm$\textbf{0. 21} & \textbf{89. 39}$\pm$\textbf{0. 13} \\
        KLSANet \cite{sun2024klsanet} & Conv4-64 & 52.23$\pm$0.56 & 70.45$\pm$0. 37 & 54,71$\pm$0. 77 & 78.47$\pm$0. 57 & 66. 70$\pm$0. 82 & 83.63$\pm$0. 28 \\
                
        ours & Conv4-64 & \textbf{55. 72}$\pm$\textbf{0. 72} & \textbf{72. 76}$\pm$\textbf{0. 48} & 56. 95$\pm$0. 66 & 81. 91$\pm$0. 34 & 65. 45$\pm$0. 67 & 79. 63$\pm$0. 33 \\

         LMPNet\cite{huang2021local} & ResNet-12 & $61.89 \pm 0.10$ & $68.21 \pm 0.11$ & $68.31 \pm 0.45$ & $80.27 \pm 0.23$ & $65.59 \pm 0.68$ & $68.19 \pm 0.23$ \\
         KLSANet\cite{sun2024klsanet} & ResNet-12 & $64.43 \pm 0.81$ & $81.07 \pm 0.31$ & $74.43 \pm 0.76$ & $87.84 \pm 0.45$ & $74.94 \pm 0.43$ & $88.92 \pm 0.41$ \\
         ours & ResNet-12 & $\textbf{73.63} \pm \textbf{0.69}$ & $\textbf{84.38} \pm \textbf{0.44}$ & $\textbf{85.16} \pm \textbf{0.60}$ & $\textbf{94.32} \pm \textbf{0.28}$ & $\textbf{79.79} \pm \textbf{0.64}$ & $\textbf{91.18} \pm \textbf{0.35}$ \\

        \bottomrule
    \end{tabular}}

\end{table*}

\begin{table*}[!t]
   \caption{Cross-domain performance comparison of the proposed FAFD-LDWR u with state-of-the-art methods on miniImageNett→CUB setting. ‘–’: not reported.}
   \label{tab:cross_domian_results}
    \centering
\begin{tabular}{lllr}
 \hline Method & Backbone & miniImageNet $\rightarrow$ CUB \\
\cline { 3 - 4 } & & 5 -way 1-shot & 5-way 5-shot \\
\hline 
ProtoNet \cite{snell2017prototypical}& Conv4-64 & $39.39 \pm 0.68$ & $56.06 \pm 0.66$ \\
RelationNet \cite{sung2018learning}& Conv4-64 & $39.30 \pm 0.66$ & $53.44 \pm 0.64$ \\
BDLA \cite{zheng2023bdla}& Conv4-64 & $40.40 \pm 0.76$ & $58.23 \pm 0.72$ \\
DLDA  \cite{song2024learning}& Conv4-64 & $41.36 \pm 0.74$ & $\textbf{60.02} \pm \textbf{0.71}$ \\
FAFD-LDWR(ours)& Conv4-64 & $\textbf{42.98} \pm \textbf{0.59}$ & $59.43 \pm 0.53$ \\

Fine-tuning \cite{sun2021explanation} & ResNet-10 & $41.98 \pm 0.41$ & $58.75 \pm 0.36$ \\
RelationNet \cite{sung2018learning} & ResNet-18 & $42.91 \pm 0.78$ & $57.71 \pm 0.73$ \\
LRP-RN \cite{hu2022adversarial} & ResNet-10 & $42.44 \pm 0.41$ & $59.30 \pm 0.40$ \\
MN+AFA \cite{chen2021meta} & ResNet-10 & $41.02 \pm 0.40$ & $59.46 \pm 0.40$ \\
PDN-PAS \cite{chen2023few} & ResNet-18 & $42.41 \pm 0.84$ & $61.25 \pm 0.86$ \\
Baseline++ \cite{fu2021meta} & ResNet-18 & $43.04 \pm 0.60$ & $62.04 \pm 0.76$ \\
Baseline \cite{fu2021meta} & ResNet-18 & - & $65.57 \pm 0.70$ \\
MatchingNet \cite{vinyals2016matching} & ResNet-18 & $45.59 \pm 0.81$ & $53.07 \pm 0.74$ \\
ProtoNet \cite{snell2017prototypical} & ResNet-18 & $45.31 \pm 0.78$ & $62.02 \pm 0.70$ \\
GNN \cite{satorras2018few} & ResNet-10 & $45.69 \pm 0.68$ & $62.25 \pm 0.65$ \\
GNN+FT \cite{tseng2020cross} & ResNet-10 & $47.47 \pm 0.75$ & $66.98 \pm 0.68$ \\
FDMixup \cite{gao2024few} & ResNet-10 & $46.38 \pm 0.68$ & $64.71 \pm 0.68$ \\
MIFN \cite{gao2024few} & ResNet-12 & $48.21 \pm 0.60$ & $65.33 \pm 0.54$ \\
KLSANet \cite{sun2024klsanet} & ResNet-12 & $48.16 \pm 0.64$ & $67.25\pm 0.61$ \\
FAFD-LDWR(ours) & ResNet-12 & $\textbf{48.64} \pm \textbf{0.42}$ & $\textbf{67.36}\pm \textbf{0.55}$ \\
\hline
\end{tabular}
\end{table*}


\begin{table}[!t]
   \caption{The influence of using local descriptor neighborhood representation(NR).
}
   \label{tab:neighbor}
    \centering
\begin{tabular}{lcll}
 \hline Method & NR &  CUB&  CUB \\
\cline { 3 - 4 } & & 5 -way 1-shot & 5-way 5-shot \\
\hline 

FAFD-LDWR & w/o & $79.63 \pm 0.54$ & $90.32\pm 0.61$ \\

FAFD-LDWR & w & $\textbf{79.79} \pm \textbf{0.64}$ & $\textbf{91.18}\pm \textbf{0.35}$ \\
\hline
\end{tabular}
\end{table}

\begin{table}[!t]
\centering
\caption{Comparison of using L2 normalization (L2) and cross normalization (CN) for the DN4 and FAFD-LDWR methods on the CUB dataset. `NM' stands for normalization method.}
\label{tab:comparison_N}
\begin{tabular}{lcll}
 \hline Method & NM &  CUB&  CUB \\
\cline { 3 - 4 } & & 5 -way 1-shot & 5-way 5-shot \\
\hline 
DN4\cite{li2019revisiting} & L2 & $46.84 \pm 0.81$ & $74.92\pm 0.64$ \\
DN4\cite{li2019revisiting} & CN & $48.60 \pm 0.62$ & $75.71\pm 0.49$ \\
FAFD-LDWR & L2 & $77.94 \pm 0.50$ & $88.37\pm 0.62$ \\

FAFD-LDWR & CN & $\textbf{79.79} \pm \textbf{0.64}$ & $\textbf{91.18}\pm \textbf{0.35}$ \\
\hline
\end{tabular}
\end{table}


\subsection{Experimental Results}
\subsubsection{General Few-Shot Classification}

To validate the effectiveness of our proposed FAFD-LDWR method, we compare it with 13 state-of-the-art few-shot classification methods on three fine-grained datasets, as summarized in Table \ref{tab:results1}.

Using the Conv-4 backbone, FAFD-LDWR shows a significant improvement in accuracy over the DN4 \cite{li2019revisiting} method, which does not process local descriptors. This highlights how poor local descriptor representations can degrade classification performance in fine-grained image classification scenarios.

However, FAFD-LDWR with the Conv-4 backbone does not show a significant advantage over recent methods like DLDA \cite{song2024learning}, MADN4 \cite{li2020more}, BDLA \cite{zheng2023bdla}, and KLSANet \cite{sun2024klsanet} across most settings. This is because Conv-4 extracts only 441 local descriptors per image, and our adaptive threshold filtering performs better with more descriptors. Therefore, we conducted further experiments using ResNet-12 as the backbone to extract more detailed local descriptors. Notably, FAFD-LDWR with ResNet-12 outperforms all compared methods across most settings on the three datasets. The reduction in noisy local features allows for a more accurate depiction of discriminative regions, resulting in significant improvements over other methods.

\subsubsection{Cross-domain Few-Shot Classification}

To evaluate the cross-domain generalization of FAFD-LDWR, we conducted experiments under the miniImageNet→CUB setting (see Table \ref{tab:cross_domian_results}) and compared it with state-of-the-art methods. The model is trained on 64 base classes of miniImageNet and evaluated on 50 novel classes in the CUB test set. Our FAFD-LDWR demonstrates significant advantages in this cross-domain scenario. Using the Conv-4 backbone, FAFD-LDWR maintains a lead in both 5-way 1-shot and 5-way 5-shot settings compared to methods that also focus on enhancing semantic alignment of local descriptors, such as BDLA \cite{zheng2023bdla} and DLDA \cite{song2024learning}.

With the ResNet-12 backbone, FAFD-LDWR achieves an accuracy of 48.64\% in the 5-way 1-shot setting and 67.36\% in the 5-way 5-shot setting. This not only surpasses classical few-shot methods like MatchingNet \cite{vinyals2016matching}, ProtoNet \cite{snell2017prototypical}, RelationNet \cite{sung2018learning}, and GNN, but also maintains a lead over methods specifically tailored for cross-domain scenarios, such as Finetuning \cite{sun2021explanation}, LRP-RN \cite{hu2022adversarial}, MN+AFA \cite{chen2021meta}, Baseline \cite{fu2021meta}, Baseline++ \cite{fu2021meta}, GNN+FT \cite{tseng2020cross}, and FDMixup \cite{gao2024few}.

\begin{figure}[ht]
\centering
\includegraphics[scale=0.3]{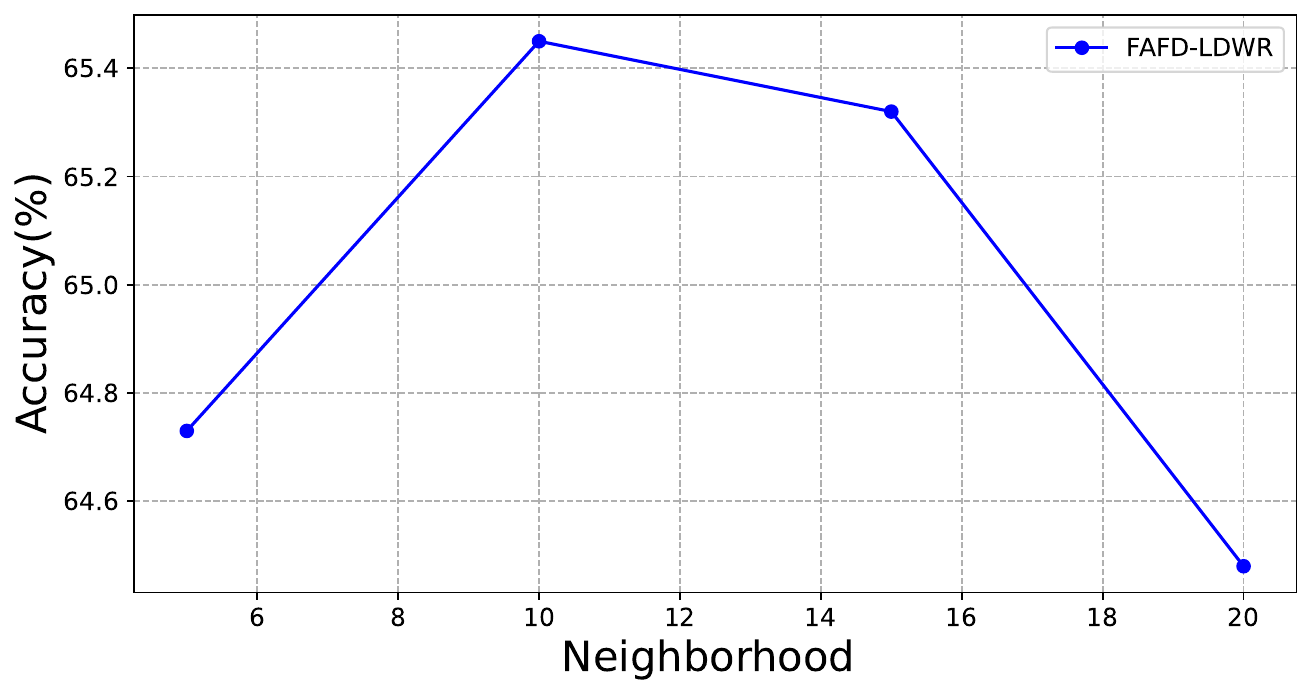}
\caption{Accuracy As A Function Of Local Descriptor Neighborhood  Representation $k$ Value.}
\label{linyu}
\end{figure}
\subsection{Ablation Studies}
\subsubsection{Impact of Local Descriptor Neighborhood Representation}
This paper proposes the innovative FAFD-LDWR method, which utilizes the neighborhood representation of local descriptors instead of directly using the local descriptors to enhance classification performance. In this section, we investigate the effect of neighborhood representation on smoothing local noise by comparing the experimental accuracy of using neighborhood representation versus directly using local descriptors, as shown in Table \ref{tab:neighbor}. Here, "w/" and "w/o" denote the usage and non-usage of neighborhood representation, respectively. The results demonstrate that calculating the mean of neighbors as a new representation can smooth local noise, thereby improving feature stability and robustness.
\subsubsection{Impact of Cross Normalization}
In our work, we innovatively introduce cross normalization for local descriptors in few-shot learning to preserve the discriminative information of local descriptors, facilitating subsequent dynamic filtering of local descriptors. A comparison between cross normalization and the commonly used L2 normalization in previous works can be found in Supplementary Section A. As shown in Table \ref{tab:comparison_N}, we conducted comparative experiments using cross normalization and L2 normalization on both the DN4 method, which does not involve any post-processing of local descriptors, and our FAFD-LDWR method. The experimental results demonstrate the effectiveness of cross normalization in improving classification accuracy.

\subsubsection{Impact of the Number of Neighbors Used in Local Descriptor Neighborhood Representation}
The proposed FAFD-LDWR method utilizes the neighborhood representation of local descriptors instead of directly using the local descriptors to enhance classification performance. In this section, we examine the impact of the number of neighbors selected for the neighborhood representation on the experimental results. The results, as shown in Figure \ref{linyu}, demonstrate that calculating the mean of neighbors as a new representation can smooth local noise, thereby improving feature stability and robustness. However, the number of neighbors selected is crucial; more neighbors do not necessarily lead to better performance. The experimental results indicate that the optimal number of neighbors is 10. This is because an excessively large neighborhood may include noise information irrelevant to the local descriptor.

\subsubsection{Impact of Different $k$ Values in $k$-NN Classifier on Experimental Results}

Detailed analysis can be found in Supplementary Sections E.

\subsubsection{Time Complexity Analysis}
We also performed a time complexity analysis, demonstrating the effectiveness of our method without increasing time complexity. Detailed analysis can be found in Supplementary Sections D.

\section{Conclusion}
In this study, we propose a  effective FAFD-LDWR method to enhance the performance of few-shot learning.

This approach enables the feature extractor to effectively focus on local descriptors relevant to the image class, thereby reducing the interference of class-irrelevant information.

Our dynamically weighted local descriptor module focuses on class-relevant key information, enhancing image representation and reducing the impact of irrelevant regions. This improves classification accuracy by filtering out irrelevant background descriptors. The method remains simple and lightweight, introducing no additional learnable parameters and maintaining consistency between training and testing phases.

The proposed method is expected to work in other data
modalities such as medical images and text data, which will
be investigated in future work

\bibliographystyle{unsrtnat}
\bibliography{references}  






\end{document}